\documentclass[lettersize,journal]{IEEEtran}
\usepackage{amsfonts}
\usepackage{mathtools}
\usepackage{algorithmic}
\usepackage{algorithm}
\usepackage{array}
\usepackage[caption=false,font=normalsize,labelfont=sf,textfont=sf]{subfig}
\usepackage{textcomp}
\usepackage{stfloats}
\usepackage{url}
\usepackage{verbatim}
\usepackage{graphicx}
\usepackage{cite}
\usepackage{makecell}
\usepackage{multirow}
\usepackage{tablefootnote}
\usepackage{arydshln}
\usepackage{xcolor}
\usepackage{enumitem}
\usepackage{xurl}
\usepackage{caption}
\usepackage{subcaption}
\usepackage[capitalize]{cleveref}
\crefname{section}{Sec.}{Secs.}
\Crefname{section}{Section}{Sections}
\Crefname{table}{Table}{Tables}
\crefname{table}{Tab.}{Tabs.}
\begin{document}

\title{StyleInject: Parameter Efficient Tuning of Text-to-Image Diffusion Models}

\author{Mohan~Zhou,
        ~Yalong~Bai,
        ~Qing~Yang,
        ~Tiejun~Zhao}

\markboth{Journal of \LaTeX\ Class Files,~Vol.~14, No.~8, August~2021}%
{Shell \MakeLowercase{\textit{et al.}}: A Sample Article Using IEEEtran.cls for IEEE Journals}


\maketitle

\begin{abstract}
The ability to fine-tune generative models for text-to-image generation tasks is crucial, particularly when facing the complexity involved in accurately interpreting and visualizing textual inputs. While LoRA is efficient for language model adaptation, it often falls short in text-to-image tasks due to the intricate demands of image generation, such as accommodating a broad spectrum of styles and nuances. To bridge this gap, we introduce StyleInject, a specialized fine-tuning approach tailored for text-to-image models. StyleInject comprises multiple parallel low-rank parameter matrices, maintaining the diversity of visual features. It dynamically adapts to varying styles by adjusting the variance of visual features based on the characteristics of the input signal. This approach significantly minimizes the impact on the original model's text-image alignment capabilities while adeptly adapting to various styles in transfer learning. StyleInject proves particularly effective in learning from and enhancing a range of advanced, community-fine-tuned generative models. Our comprehensive experiments, including both small-sample and large-scale data fine-tuning as well as base model distillation, show that StyleInject surpasses traditional LoRA in both text-image semantic consistency and human preference evaluation, all while ensuring greater parameter efficiency.
\end{abstract}


\section{Introduction}
\IEEEPARstart{T}{he} field of text-to-image generation, a key component of AI-driven content creation, is experiencing significant advancements. Central to this progress are extensive training sets~\cite{schuhmann2022laion} and large-scale generative models~\cite{saharia2022photorealistic,mishkin2022risks,rombach2022high,nichol2021glide}, which have shown remarkable prowess in interpreting and converting textual descriptions into images. Among these, Stable Diffusion models (SDMs)~\cite{rombach2022high,podell2023sdxl} are particularly noteworthy as leading open-source options. Their exceptional capabilities, coupled with their efficiency in performing high-quality inference in low-dimensional latent spaces, have been a foundational technology in various text-guided vision applications. 

One of the cornerstones of large-scale generative model revolution has been the capability to fine-tune these models, tailoring them to specific tasks or styles. Traditionally, fine-tuning large pre-trained models has been the go-to strategy for adapting to specific tasks or domains. However, this approach poses significant challenges in efficiency, particularly as model sizes increase. LoRA (Low-Rank Adaptation)~\cite{hu2021lora}, a method proposed to address these challenges, has emerged as a frontrunner in fine-tuning pretrained large language models~\cite{liu2019roberta,brown2020language}. Typically, LoRA reduces the trainable parameter count significantly by introducing low-rank matrices into each layer of a Transformer model, thus maintaining the pre-trained weights frozen. Its efficiency and reduced memory overhead have motivated community also to explore its potential in the realm of image generation. 

However, unlike text that conveys a single narrative thread, a textual description can be interpreted into a vast array of visual representations, introducing a higher degree of uncertainty in the generated content. Traditional adaptation techniques like LoRA employ a linear approach to fine-tuning, which involves augmenting the original model with additional low-rank matrices. While effective in some contexts, this method can inadvertently lead the model to favor certain visual styles, limiting its versatility on image generation. This predisposition may compromise the precise alignment between text and the corresponding images. In practice, to counteract this tendency and preserve a broader style range, the community often resorts to early stopping and cherry-pick checkpoint during training of text-to-image models with LoRA. However, as a suboptimal solution, it restricts the models from generating images of singular styles or with limited semantic variance, thus not fully exploiting the potential diversities.

In this work, we revisited the LoRA fine-tuning mechanism and evolved it into a method apt for visual content generation – StyleInject. First, to accommodate the diversity and uncertainty of visual features, we propose a parallel representation of ``intrinsic rank'' to cater to the variety in visual features. Building on this, we utilize dynamic neural networks for instance-wise feature adaptation. Finally, to minimize the potential disruption of the original model's text-image alignment capability due to direct introduction of linear adaptation, we design a novel style transfer module. Influenced by traditional style transfer algorithms~\cite{huang2017arbitrary,nam2018batch,karras2019style,DBLP:journals/tmm/WangLWZF24,DBLP:journals/tmm/BaiXZD24}, this module deconstructs the visual features outputted by the pretrained model, and then modifies and injects new variances to facilitate a comprehensive stylistic transformation of visual features. Overall, this approach aims to tune text-to-image models while preserving the original model's capability for text-image alignment.

Furthermore, we extend the application of StyleInject, beyond the traditional data-driven fine-tuning tasks to encompass model distillation as well. This approach enables us to leverage the collective insights and improvements made by the broader open-source Stable Diffusion models community, further enhancing the robustness and versatility of our method. Moreover, to rigorously test the capability of StyleInject in maintaining text-image alignment, we introduce an innovative cross-lingual text-to-image model distillation task. The results from these diverse experimental scenarios collectively underscore the effectiveness, stability, and efficiency of our proposed fine-tuning method. The experimental outcomes demonstrate that StyleInject not only adeptly adapts to varying visual styles but also preserves the nuanced relationship between text and image, a critical aspect in the realm of generative models.

The main contributions are summarized as follows:
\begin{itemize}
    \item We proposed a novel adaptation approach for text-to-image generation, uniquely adapting style variance within image features. This method enhances semantic accuracy and style adaptability, addressing the limitations of existing linear adaptation methods.
    \item A practical text-to-image model distillation task is introduced to test the capability of StyleInject while maintaining semantic coherence. By successfully maintaining high fidelity in semantic understanding despite linguistic variances, our method shows its potential to enhance and enrich the multilingual capabilities of generative models.
    \item Our extensive experiments demonstrate the superiority of StyleInject over traditional methods in effectiveness, stability, and efficiency. Additionally, by applying our method to community-fine-tuned models, we showcase its capacity to enhance pre-existing models and contribute back to the broader AI community.
\end{itemize}

\section{Related Work}

\noindent\textbf{Text-to-Image stable diffusion models}\quad In the realm of generative models, diffusion models~\cite{ho2020denoising,song2020denoising} have emerged as a groundbreaking advancement to GAN-based models~\cite{goodfellow2014generative,DBLP:journals/tmm/XiaoB24,DBLP:journals/tmm/ZhengBLWYS23,DBLP:journals/tmm/HouZLS23,DBLP:journals/tmm/YinCP22,DBLP:journals/tmm/TanLYL22,DBLP:journals/tmm/ZhangLXYLC22}. Leveraging the principles of diffusion processes, these models have shown remarkable success in generating high-quality, diverse images from textual descriptions. Stable diffusion models (SDMs)\cite{rombach2022high,podell2023sdxl}  perform diffusion modeling in a 64×64 latent space constructed through a pixel-space autoencoder. Due to the low-cost computation in latent space and its open-access nature, SDMs are rapidly gaining popularity across numerous downstream tasks, such as image editing\cite{brooks2023instructpix2pix,hertz2022prompt}, subject-driven image generation~\cite{ruiz2023dreambooth}, and controllable image generation~\cite{Zhang_2023_ICCV,DBLP:journals/tmm/CaoCHZCW24}. Concurrently, this has led to the emergence of thousands of community-fine-tuned models. A significant advantage of these community models is their ability to generate images of certain specific styles or content with enhanced quality, thanks to fine-tuning on high-quality data. The fine-tuning method proposed in this paper attempts to utilize both data-driven and data-free approach to transfer the image styles from the most widely acclaimed foundational model checkpoints in the community.

\noindent\textbf{Parameter efficient tuning}\quad The fine-tuning of foundational models, especially large-scale language models, has emerged as a pivotal factor in advancing AI capabilities. Techniques such as LoRA (Low-Rank Adaptation)\cite{hu2021lora} have proven efficient in adapting these models with significantly fewer trainable parameters, striking a balance between performance and computational efficiency. Furthermore, there are numerous variants based on the LoRA model, each addressing specific challenges. For instance, KronA~\cite{edalati2022krona} introduces a Kronecker product-based adapter module for efficient fine-tuning, AdaLoRA~\cite{zhang2023adaptive} adjusts the rank of incremental matrices to control parameter budget, DyLoRA~\cite{valipour2022dylora} trains LoRA blocks for a range of ranks instead of a single rank by sorting the representation learned by the adapter module at different ranks during training. Nevertheless, these methods are predominantly utilized for fine-tuning large-scale language models. When applied to text-to-image tasks, which demand an understanding of complex and varied image features, these methods often fail to account for the considerable variation in the importance of weight matrices across different visual styles and semantics. This oversight underscores the need for more nuanced and context-sensitive fine-tuning approaches in the realm of multimodal AI applications.

\noindent\textbf{Model distillation for SDMs}\quad Model distillation is a technique employed to transfer knowledge from larger, more complex models to smaller, more efficient ones. In the realm of stable diffusion models (SDMs), distillation is key to enhancing their deployability and operational efficiency~\cite{meng2023distillation}. By distilling knowledge from large, pre-trained generative models into more compact forms, we can preserve much of their generative capacity while reducing computational demands~\cite{kim2023architectural,Kim_2023_ICMLW}. In this paper, we explore another application of model distillation: the distillation of image styles. This process involves learning the characteristics of high-quality finetuned models from the SDMs community and transferring them back to enhance the original foundational models, while ensuring the preservation of the original model's text-image alignment abilities.

\section{Method}
\subsection{Revisiting Low-Rank Adaptation}

Low-Rank Adaptation (LoRA)~\cite{hu2021lora} initially emerged as an efficient and effective model fine-tuning method for big language models. Subsequently, it gained widespread traction in the domain of diffusion models, wherein its utilization extends to the incorporation of diverse stylistic elements and human entities, while even serving to alleviate visual artifacts. 

Instead of exhaustive fine-tuning the full parameters of a given diffusion model, it integrates trainable lightweight rank decomposition matrices within each Attention layer. This strategy significantly diminishes the resource costs while concurrently expanding the scope of model adaptability and potential functionalities for downstream tasks.

Consider a pre-trained weight matrix \( W_0 \in \mathbb{R}^{d \times k} \) in a dense layer, output $h=W_0x$ for an input $x$. LoRA adapts this weight matrix using low-rank matrix decomposition during the adaptation process. The adapted weight matrix \( W \) is represented as $W = W_0 + BA$, where \( B \in \mathbb{R}^{d \times r} \) and \( A \in \mathbb{R}^{r \times k} \) are low-rank matrices, where \( r \) is the chosen rank. The rank \( r \), significantly smaller than \( d \) and \( k \), controls the model's adaptation flexibility. The output \( h^{*} \) for an input \( x \) in the modified layer with LoRA is computed by:

\begin{equation}
    h^{*} = h + \Delta h = W_0x + BAx.
\end{equation}

In this reparametrization, only \( A \) and \( B \) are updated during training, significantly cutting down the computational load compared to full model fine-tuning. The majority of parameters in \( W_0 \) are kept unchanged, maintaining the foundational knowledge of the pre-trained model.

LoRA's approach is beneficial when adapting a pre-trained model to multiple tasks or sceneratios. By changing only \( A \) and \( B \), the model can swiftly switch tasks while relying on a shared base of pre-trained weights.

\subsection{Our Method: StyleInject}

\begin{figure}
    \centering
    \includegraphics[width=\linewidth,page=1]{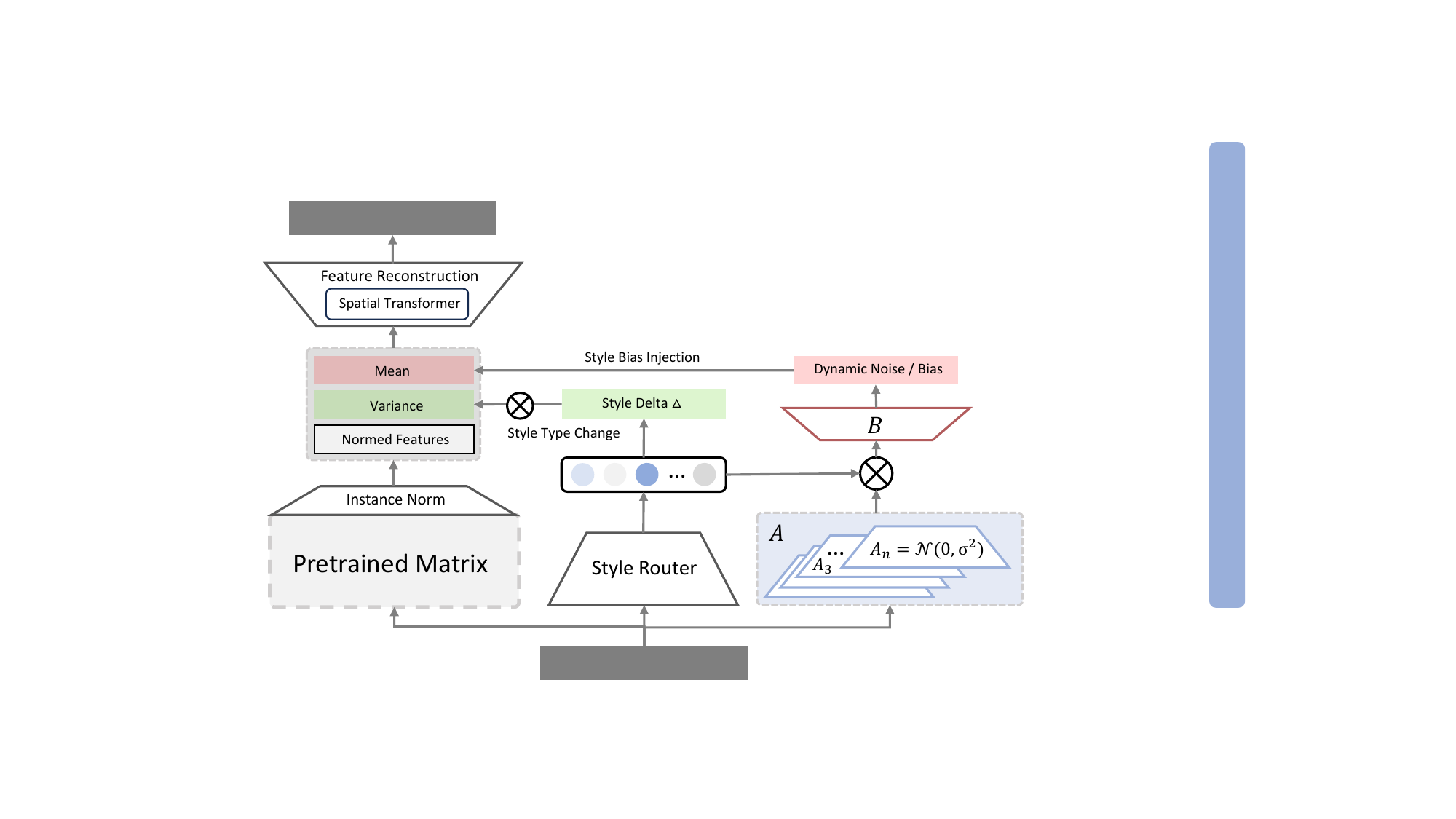}
    \caption{Workflow of the StyleInject method. Input \( x \) passes through a pretrained frozen matrix \( W_0 \) to produce features \( h \), subsequently normalized via AdaIN. A style router generates a style probability vector \( \textbf{s} \), guiding the variance adjustment through \( f_\textbf{s} \). The final visual feature \( h^* \) results from the integration of normalized features $\hat{h}$, adjusted variance $\hat{\sigma}(h)$, preserved mean $\mu(h)$, and adjustment term $\Delta h$ from dynamic multi-style adaptation, ensuring better adaptive visual features.}
    \label{fig:method}
\end{figure}

In this section, we introduce our novel parameter-efficient tuning method for text-to-image generation tasks. As illustrated in \cref{fig:method}, the StyleInject framework comprises two primary components: dynamic multi-style adaptation network, and style transfer via AdaIN.

\subsubsection{Dynamic multi-style adaptation}


In enhancing the multi-style adaptation capability of StyleInject, our approach introduces $n$ sets of down-ranked matrices $\textbf{A}=\left\{A_1, A_2,..., A_n\right\}$, combined with a shared up-ranked matrix $B$. In this way, we enable the low-rank feature space to handle diverse visual styles without creating separate spaces for each. This unified approach ensures coherence and fluidity in transitioning between styles. Moreover, the shared matrix approach avoids the parameter increase that would result from multiple $B$ matrices, which is crucial for maintaining the model's efficiency. In essence, this configuration is meticulously crafted to efficiently differentiate between visual features across a multitude of styles within a confined low-dimensional space.

The core of our adaptation mechanism is the \textit{style router}, a dynamic component that maps instance-wise visual features into an \( n \)-dimensional probability vector \( \mathbf{s}=\{s_1,s_2,...,s_n\} \). To ensure the overall parameter efficiency of the adaptation network, we employed a simple fully-connected layer (for Linear layer adaptation) or a convolutional layer of kernal size of (1x1) following feature-map summation (for Conv-2D layer adaptation) as a simple router network, the final $\mathbf{s}$ is normalized by softmax. The vector $\mathbf{s}$ plays a pivotal role in the adaptation process, dynamically guiding the selection of appropriate style combinations from \( \textbf{A} \). By doing so, the style router enables an input-specific tuning of the model, which is a hallmark of dynamic neural networks, resulting in enhancing the model's overall flexibility and expressiveness. The contribution of each style to the final adapted feature is determined by
\begin{equation}
\Delta h=B\sum_{i=1}^n s_iA_i(x),
\end{equation}
which combines the influences of all style-specific adaptations into a single modification term.

This dynamic multi-style adaptation not only caters to the diverse range of visual styles inherent in text-to-image tasks but also ensures that each generated image is a distinct and accurate reflection of the intended style.
\subsubsection{Style transfer via AdaIN}

In StyleInject, the technique of Adaptive Instance Normalization (AdaIN)~\cite{huang2017arbitrary} plays a pivotal role. It is originally developed for style transfer in images, and adept at aligning the content of one image with the style of another. In our context, it is employed to harmonize text-driven image generation with varying styles.

The process begins with the input features $h=W_0x$, transformed by the pretrained matrix. These features undergo AdaIN, which normalizes each feature channel separately. The normalization is achieved by first computing the mean $\mu(h)$ and variance $\sigma(h)$ of each channel:
\begin{equation}
\hat{h} = \frac{h-\mu(h)}{\sigma(h)}.
\end{equation}
This normalization strips the original features of their original style-specific characteristics, creating a neutral canvas to impose new styles.

Next, the dynamic style adaptation network outputs a result processed through a simple one-layer hypernetwork~\cite{ha2016hypernetworks}, generating style transfer parameters $f_{\mathbf{s}}$. These parameters are crucial as they allow us to recalibrate the variance of the normalized features, since the variance of initialization noise map~\cite{mao2023guided} or middle visual feature~\cite{yu2023constructing} has been demonstrated with deep influences on the content and style of the generated images. By modifying the original variance $\sigma(h)$ with $f_{\mathbf{s}}$, we obtain a new variance $\hat{\sigma}(h)=f_{\mathbf{s}}(\sigma(h))$. This step is where the actual `style injection' occurs, as the recalibrated variance imparts new stylistic attributes to the features.

Finally, to maintain the semantic integrity of the generated images, we employ two critical strategies:
\begin{enumerate}
    \item \textit{Initialization of Style Transfer Parameters}: We initialize $f_{\mathbf{s}}(\sigma(h))=\sigma(h)$. This initialization mirrors the approach used in LoRA and ControlNet, which minimizes initial disruption to the model's feature representation. It ensures that the original semantic content is preserved while the style is being adapted.
    \item \textit{Preservation and Reconstruction of Features}: After AdaIN decomposition, we preserve the mean value $\mu(h)$. The reconstructed visual features then combine this mean with the modified variance $\hat{\sigma}(h)$ and the normalized features. The final visual representation is obtained by:
    \begin{equation}
    h^{*} = \hat{h}f_{\mathbf{s}}\left(\sigma(h)\right) + \mu(h) + \Delta h.
    \end{equation}
    This combination ensures that while the style is dynamically adapted, the underlying content, driven by the text input, remains semantically coherent.
\end{enumerate}

Through these mechanisms, StyleInject leverages AdaIN not just as a tool for style transfer, but as a means to dynamically and coherently blend various styles with the content generated from textual descriptions. This approach ensures that the generated images are not only stylistically diverse but also maintain a strong alignment with the textual input, a critical aspect in text-to-image generation tasks.
\section{Distillation from Community SDMs}\label{sec:method_distill}
As mentioned previously, the open-source community around Stable Diffusion models (SDM) harbors a plethora of high-quality base model checkpoints. These models are typically fine-tuned versions of SDM's open-source iterations, augmented with user-collected high-quality training datasets.

A natural progression of thought leads us to ponder how we might leverage the visual styles embedded within these SDMs, absent a high-quality image dataset. This concept aligns with a straightforward model distillation process. Taking this a step further, imagine if we could distill styles from sophisticated, community-developed teacher models into streamlined, low-parametric adaptation networks, we could then \textit{emulate the diverse performance of various community SDMs within a single general and foundational model, achieved simply by flexibly integrating or separating these adaptation networks}. This approach not only reduces the complexity of managing multiple SDMs but also crucially minimizes the computational overhead typically associated with running multiple SDMs concurrently. It presents a more efficient paradigm, where the agility of model adaptation takes precedence without extensive resource consumption.

Considering the characteristics of SDMs, we can categorize the task of distillation from community SDMs into two scenarios: 1) where the teacher and student models use a \textit{shared text encoder}, and 2) where they employ different \textit{unshared text encoders}, with the former scenario being less complex than the latter.  

In the context of stable-diffusion-based image generation, wherein an image is represented by the latent vector $z$ and accompanied by a corresponding textual description $y$. The knowledge distillation technique, referred to `distill-sd'~\cite{kim2023architectural,Kim_2023_ICMLW,DBLP:journals/tmm/MaLXYDJ24}, is mostly trained through the optimization of the task loss $\mathcal{L}_\text{Task}$ alongside various measures of feature consistency. These measures include $\mathcal{L}_\text{OutKD}$ for aligning output-level features or $\mathcal{L}_\text{FeatKD}$ for facilitating alignment at the block-wise level. Consequently, this framework enables the student model $\epsilon_S$ to learn diverse characteristics from the teacher model $\epsilon_T$.

\begin{equation}
\begin{aligned}
    \mathcal{L}_\text{Task} &= \mathbb{E}\left[\left\Vert \epsilon - \epsilon_S(z_t, y, t) \right\Vert_2^2\right], \\
    \mathcal{L}_\text{OutKD} &= \mathbb{E}\left[\left\Vert \epsilon_T(z_t, y, t) - \epsilon_S(z_t, y, t) \right\Vert_2^2\right], \\
    \mathcal{L}_\text{FeatKD} &= \mathbb{E}\left[\sum_l\left\Vert f_T^l(z_t, y, t) - f_S^l(z_t, y, t) \right\Vert_2^2 \right],
\end{aligned}
\end{equation}

where $\epsilon$ represents the sampled noise in the diffusion process, and $f_\cdot^l$ can extract the feature map from layer $l$ for the student model or teacher model. The final objective of the distillation process is always composed of these three weighted loss tunning.
\begin{equation}
    \mathcal{L} = \mathcal{L}_\text{Task} + \lambda_\text{OutKD}\mathcal{L}_\text{OutKD} + \lambda_\text{FeatKD}\mathcal{L}_\text{FeatKD}.
\end{equation}

However, our task diverges from conventional knowledge-distillation training as the student network is already a well-trained model, and high-quality training image-text pairs are not available, making direct optimization of \(\mathcal{L}_\text{task}\) unfeasible. To foster a more comprehensive transfer of knowledge from the teacher model, we opt for distillation training over a diverse set of generic image-text data.

Consequently, for scenario 1) \textit{shared text encoder}, we utilize \(\mathcal{L}_\text{OutKD} + \mathcal{L}_\text{FeatKD}\) as our distillation training loss function. And for scenario 2) \textit{unshared text encoder}, due to the substantial differences in text feature encoding, minimizing \(\mathcal{L}_\text{FeatKD}\) risks excessive damage to the student model, potentially leading to model collapse; hence we only employ \(\mathcal{L}_\text{OutKD}\) for the distillation training loss function. In our distillation process all parameters are frozen except the injected light-weight adaptation network.

\section{Experimental Results}
To validate the efficiency of our proposed method in both fine-tuning training and model distillation tasks, we conducted comparative experiments across three distinct tasks: traditional data-driven model fine-tuning, distillation from community SDMs, and concept learning via DreamBooth. These tasks were aligned with experimental settings commonly found in text-to-image generation literature.

\begin{figure}[t]
\centering
\includegraphics[width=\linewidth,page=1]{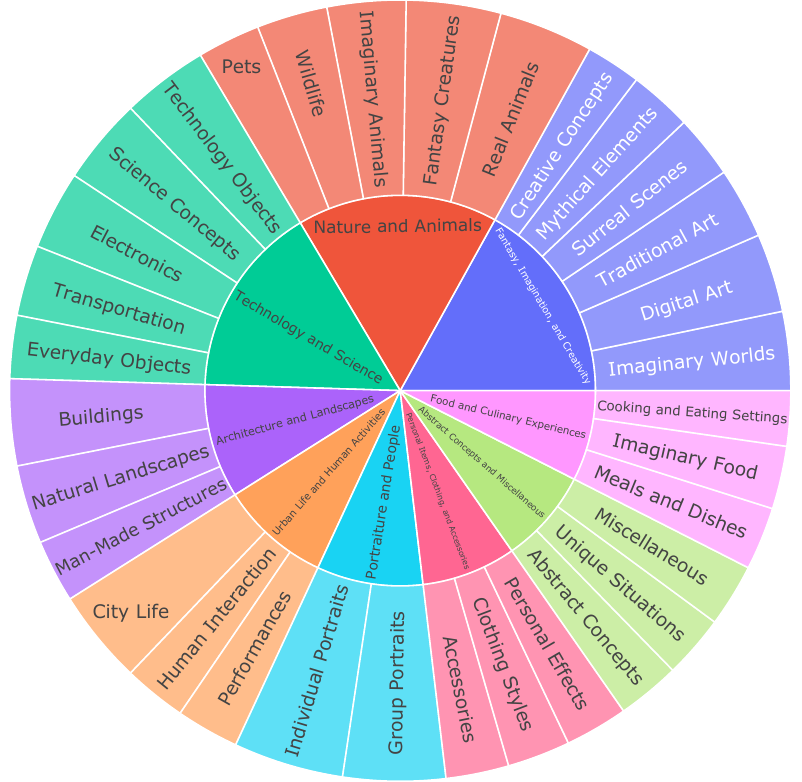}
\caption{Sunburst chart illustrating the distribution of test prompts used for model evaluation, ranging from everyday objects to imaginative worlds, ensuring diverse and thorough testing of the text-to-image generation model's capabilities.}
\label{fig:test_prompts}
\end{figure}

\textbf{Implementation Details}\quad Our experiments were based on PEFT\footnote{\url{https://github.com/huggingface/peft}}, implemented StyleInject as an additional supported method for supporting various architectures. Unless otherwise specified, we employed Stable Diffusion v1.5 (SD 1.5) as the base/student model. During fine-tuning training, StyleInject followed the hyperparameter setting of LoRA fine-tuning with a learning rate of 1e-4 and a constant learning rate schedule.

For fair comparisons, all experiments are conducted using a $512\times512$ resolution on 4 GPUs with a batch size of 32 and 2 steps of gradient accumulation. The random seed for training process was also kept consistent. Besides, we set $r=32$, $n=16$ for StyleInject and $r=512$ for LoRA, and set both {\tt lora\_alpha} to \(1.0 \times r\). This also ensures that the parameter scaling is consistent between the two methods. To best showcase the generative capabilities of the models, we utilized the Euler Ancestral Discrete Scheduler with a step size of 50 for all experiments.

Notably, some community models undergo fine-tuning alongside the CLIP text encoder during training. However, our evaluations revealed that using the original version of the SD text encoder does not significantly impact the performance of community models. To align with scenario 1 of distillation from community models (\cref{sec:method_distill}), all community SD models in our study used the text encoder from the original SD model.

\textbf{Test Prompts}\quad To thoroughly assess the performance of our text-to-image generation model, we curated a set of 200 prompts for testing. This set was meticulously compiled, combining manually constructed prompts with high-frequency prompts sourced from DiffusionDB.  These prompts covered a wide range of topics, including artistic design, natural photography, portraits and animal photos, poetry and proverbs, and geometric patterns. To facilitate effective thematic categorization, we input these prompts into GPT-4 for automatic theme clustering. The results, as depicted in \cref{fig:test_prompts}, show a well-balanced distribution of prompts across 9 major categories and 33 distinct themes. This diverse and comprehensive collection of prompts was instrumental in ensuring a holistic evaluation of the model’s capabilities in generating images across a wide range of topics and styles. For each prompt, four images were generated using four fixed random seeds, and the average score of all 800 images was used as the final quantitative result.

\textbf{Evaluation Metrics}\quad We undertake a thorough quantitative assessment of our generated images, delineated along two crucial dimensions: 

1) Alignment with the Prompt. This dimension evaluates how closely the generated image corresponds to the provided prompt. It's essential to ensure that the generated content remains faithful to the intended concept or scenario. To quantify this alignment, we leverage CLIPScore~\cite{hessel2021clipscore} (CLIP-S\footnote{\url{https://github.com/jmhessel/clipscore}}). CLIPScore offers a robust metric for measuring the semantic similarity between textual prompts and generated images, providing valuable insights into the effectiveness of our model's alignment capabilities.

2) Human Preferences. Understanding human preferences towards generated images is crucial for assessing their quality and relevance. We employ two distinct methodologies to capture human preferences effectively:
\begin{itemize}
    \item ImageReward~\cite{xu2023imagereward} (IR\footnote{\url{https://github.com/THUDM/ImageReward}}). As the first general-purpose text-to-image human preference model, ImageReward offers a groundbreaking approach to gauging subjective preferences. Trained on a substantial dataset comprising 137k pairs of expert comparisons, ImageReward encapsulates diverse human judgments and preferences, enriching our understanding of what resonates most with users.
    \item PickScore~\cite{kirstain2024pick} (PICK\footnote{\url{https://github.com/yuvalkirstain/PickScore}}). Another integral component of our assessment toolkit, PickScore represents a text-image scoring function honed through extensive training on a dataset of 500k samples. Remarkably, it has exhibited superhuman performance in predicting user preferences, surpassing conventional metrics and offering nuanced insights into the subjective appeal of generated images. For readability purposes, we adjust the raw score by subtracting 21.0.
\end{itemize}

By incorporating three multidimensional assessments, we can gain comprehensive insights into both the objective alignment of generated images with prompts and their subjective appeal to human preferences, facilitating continual refinement and enhancement of our image generation model.

\subsection{Data-driven SDM Finetuning}\label{sec:exp_datatune}
We utilized a subset\footnote{\url{https://huggingface.co/datasets/ChristophSchuhmann/improved_aesthetics_6.5plus}} of LAION-5B~\cite{schuhmann2022laion} for finetuning the SD 1.5 model. To mitigate the adverse effects of low-resolution images on overall model quality, we exclusively retained images with a minimum edge exceeding 512 pixels, resulting in a training dataset with 167k image-text pairs. 

\begin{figure}[htbp]
  \centering
    \includegraphics[width=\linewidth,page=1]{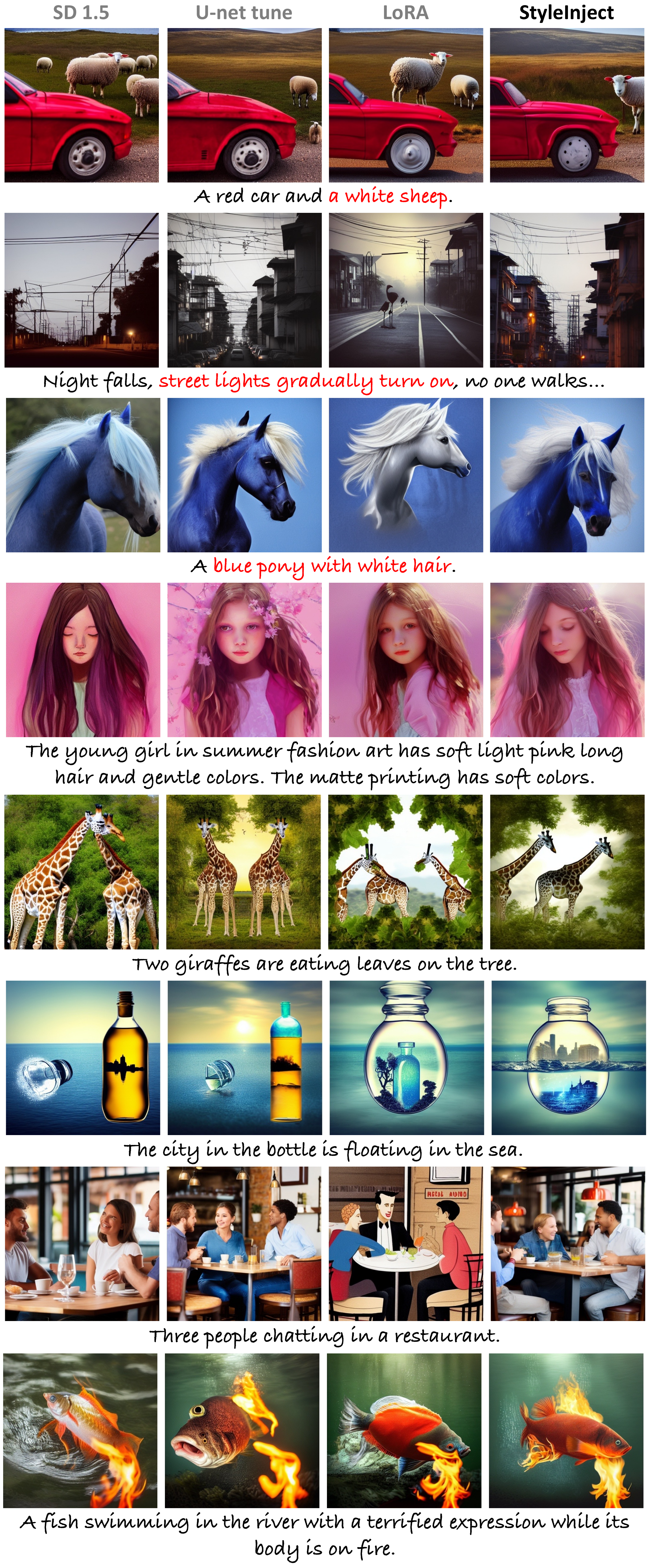}
  \caption{Examples of images generated by fine-tuned models. Each row of images is generated from the same initial noise.}
  \label{fig:datatune_visu}
\end{figure}

\cref{tab:datatune} and \cref{fig:datatune_visu} present the quantitative and qualitative results of the models post-finetuning. Our comparisons included settings where only the text encoder was frozen, finetuning all parameters of U-net (following the best practice, we set the \textit{lr} as 1e-5 for U-net tune), and using only LoRA and StyleInject for adaptation training on the transformer block's \textit{to\_q} and \textit{to\_v} linear transformations in U-net. It is important to note that our StyleInject method, being specifically designed for visual feature adaptation, is applied post the visual feature relevant \textit{to\_q} transformation. For consistency, text feature related \textit{to\_v} employs the same adaptation scheme as LoRA. 

\begin{table}[t]
\caption{Comparisons on data-driven SDM finetuning.}
\centering
\begin{tabular}{|l|c|c|c|}
\hline
 Model &  CLIP-S ($\uparrow$)    & IR ($\uparrow$)  & PICK ($\uparrow$)   \\
\hline
SD 1.5 &  0.8133 & 0.346 & 0.113 \\
\hdashline[2.5pt/2.5pt]
U-net tune & 0.8093 & 0.411 & 0.247 \\
LoRA  & 0.8079 &  0.443 & 0.183 \\
StyleInject  & \textbf{0.8149} & \textbf{0.491} & \textbf{0.306} \\
\hline
\end{tabular}
\label{tab:datatune}
\end{table}

\begin{table}[t]
\centering
    \caption{Results from ablation studies on data-driven SD 1.5 finetuning experiments.}
    \begin{tabular}{|ll|ccc|}
    \hline
     DMA & STA & CLIP-S ($\uparrow$)    & IR ($\uparrow$)  & PICK ($\uparrow$)   \\
    \Xhline{2\arrayrulewidth}
    & & 0.8097 &  0.387 & 0.203 \\
    \checkmark &  & 0.8105  & 0.484 & 0.256\\
    & \checkmark  & 0.8119 & 0.443 & 0.234\\
    \checkmark & \checkmark & \textbf{0.8149} & \textbf{0.491} & \textbf{0.306} \\
    \hline
    \end{tabular}
    \label{tab:ablation_study}
\end{table}

\subsection{Distillation on SDMs}\label{sec:exp_distillation}
We selected three of the most popular base models from the civitai site as our teacher models. These are \textbf{DreamShaper}\footnote{\url{https://huggingface.co/Lykon/DreamShaper}}, a versatile and powerful text-to-image model known for generating high-quality images with aesthetic and artistic values; \textbf{EpiCPhotoGasm}\footnote{\url{https://huggingface.co/Yntec/epiCPhotoGasm}}, specialized in realistic style image generation; and \textbf{Counterfeit-V3.0}\footnote{\url{https://huggingface.co/gsdf/Counterfeit-V3.0}}, renowned for its high-quality anime style images.

In our experiments, we employed two different student models. Considering that the text encoders of the aforementioned community models are based on the CLIP model, we used SD1.5 as one student model, aligning with scenarios 1) described in Section~\ref{sec:method_distill}. Additionally, we developed a Chinese-CLIP~\cite{chinese-clip} based ZH\_SD model, trained with commercially usable data from Midjourney, supporting Chinese prompt inputs. This ZH\_SD model served as another student model, corresponding to scenarios 2) in Section~\ref{sec:method_distill}. For the distillation task on ZH\_SD, due to the language difference between the teacher and student models, we input English prompts to the teacher model and corresponding translated Chinese prompts to the student model during training, which can be regard as a cross-lingual distillation. 
Diverging from the settings in Section~\ref{sec:exp_datatune}, where only \textit{to\_q} and \textit{to\_v} were considered, we extended the fine-tuning training to include convolutional layer parameters for better alignment with the teacher models' performance. Given that convolutional layer parameters are vital for visual features, they were fully incorporated into the StyleInject training.

\begin{figure*}[htbp]
  \centering
    \includegraphics[width=\linewidth,page=1]{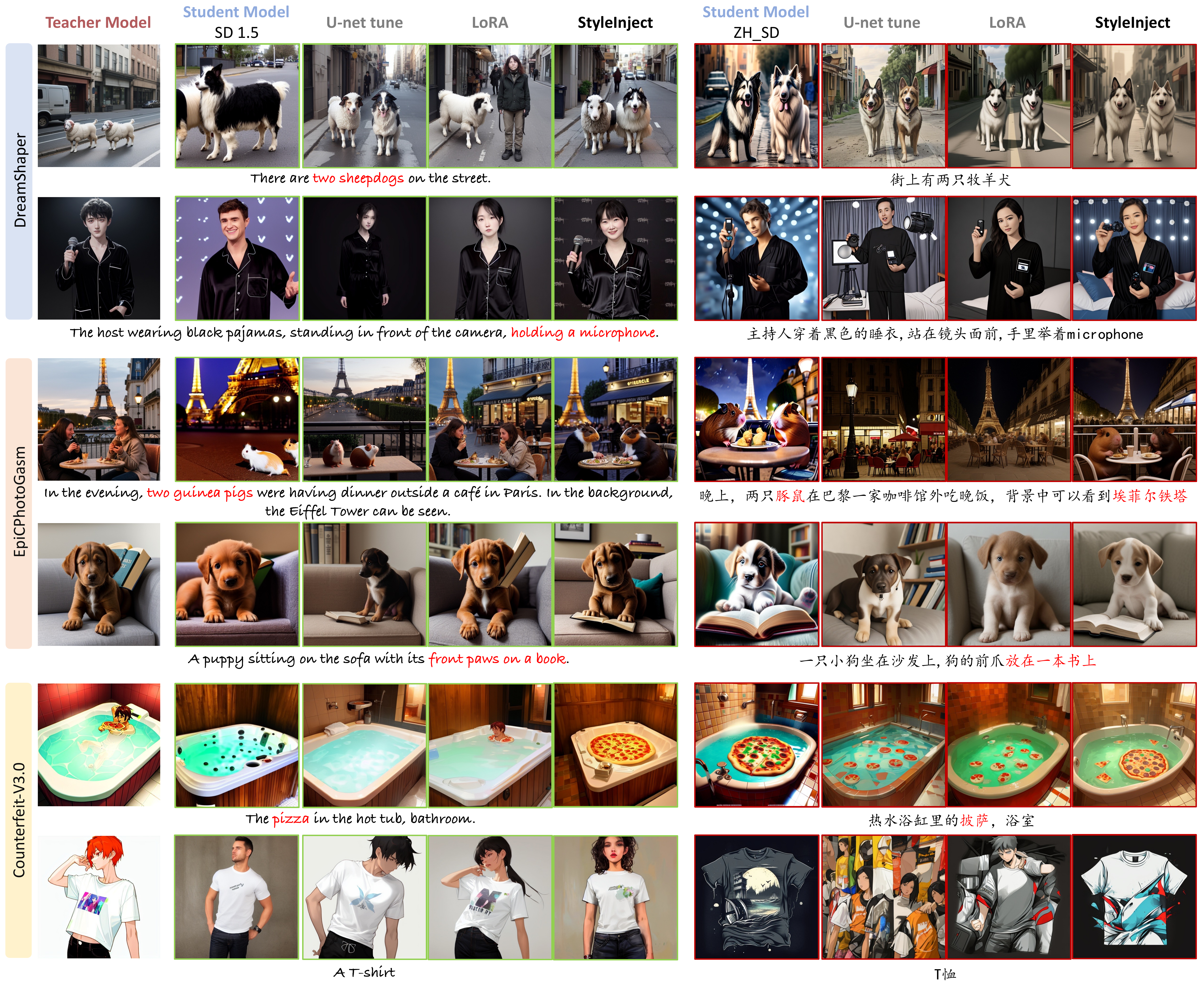}
  \caption{Visualization comparisons of SD1.5 and cross-lingual ZH\_SD distillation from community SDMs. Each row of images has been generated using the same initial noise inputs. (zooming in reveals finer details and differences in quality)}
  \label{fig:distill_visu}
\end{figure*}

\begin{figure}[htbp]
  \centering
  \includegraphics[width=\linewidth,page=2]{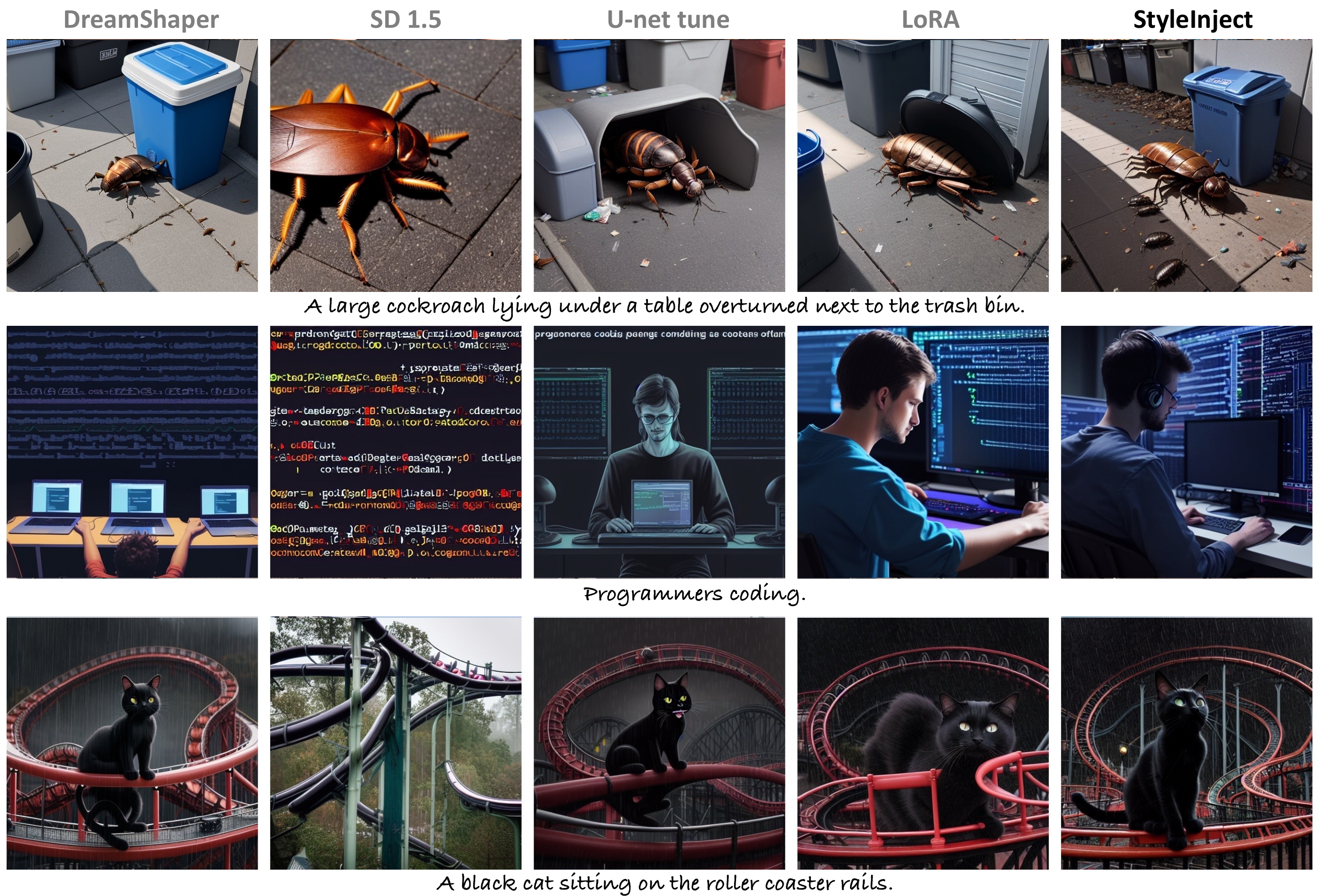}
  \caption{Examples of images generated by community model DreamShaper, original SD 1.5, and its variations distilled from DreamShaper.}
  \label{sup_fig:distill_dreamshaper}
\end{figure}

\begin{figure}[htbp]
  \centering
  \includegraphics[width=\linewidth,page=3]{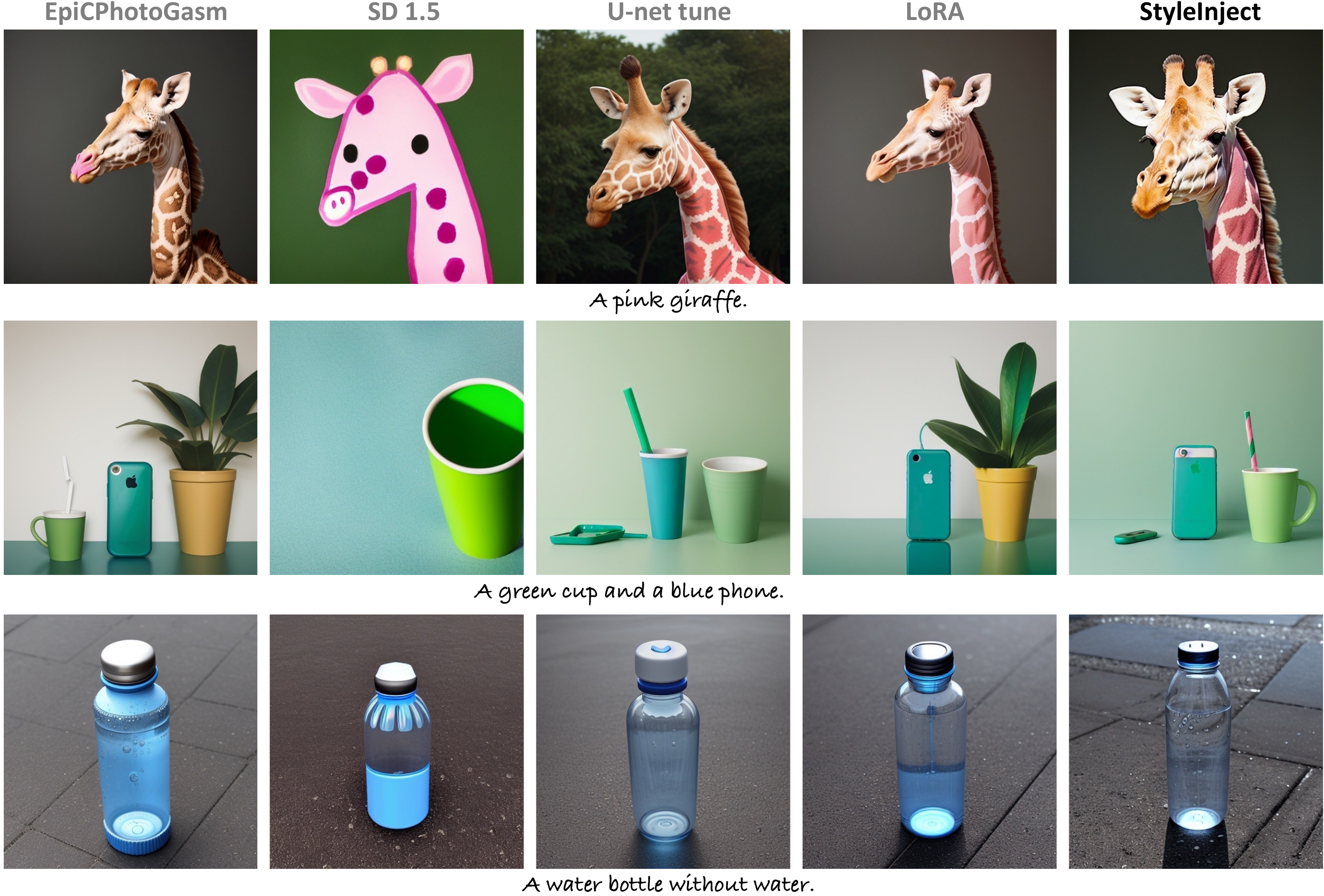}
  \caption{Examples of images generated by community model EpiCPhotoGasm, original SD 1.5 and its variations distilled from EpiCPhotoGasm.}
  \label{sup_fig:distill_epicphoto}
\end{figure}

\begin{figure}[t]
  \centering
  \includegraphics[width=\linewidth,page=4]{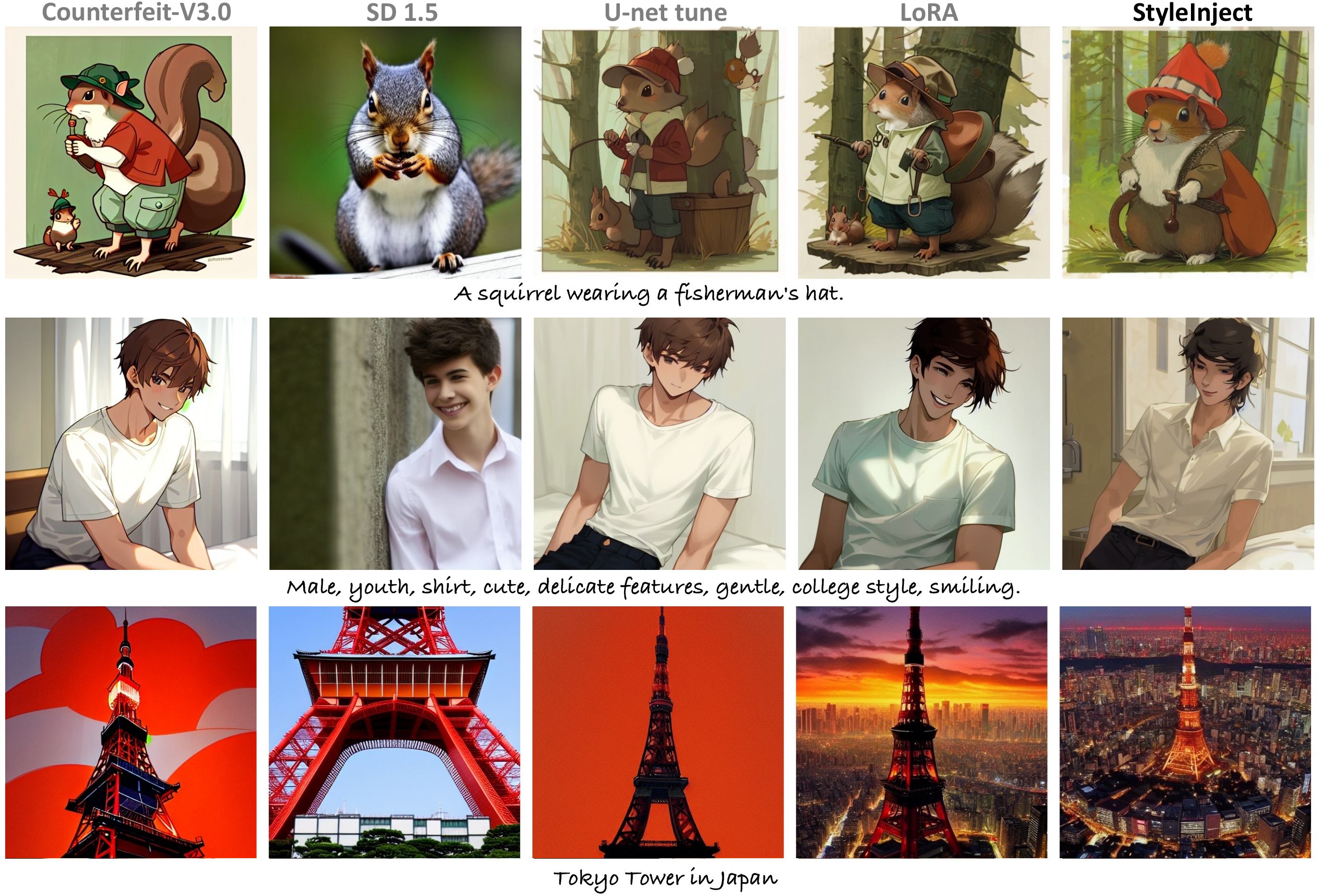}
  \caption{Examples of images generated by community model Counterfeit-V3.0, original SD 1.5 and its variations distilled from Counterfeit-V3.0.}
  \label{sup_fig:distill_cv30}
\end{figure}

Both the quantitative outcomes in \cref{tab:distill_res,tab:datatune}  and the visual comparisons in \cref{fig:distill_visu,fig:datatune_visu} demonstrate that StyleInject outperforms LoRA. This superiority is evident in more natural and richer visual details, as well as in better text-image semantic alignment. Another apparent characteristic is that community models often focus more on human-centric image generation. In other words, these models exhibit a certain semantic bias, leading them to produce human-like figures even when generating objects unrelated to humans. As seen in rows 3 and 5 of \cref{fig:distill_visu}, the straightforward linear combination mechanism of LoRA with the base model make it susceptible to such semantic biases. While the style transfer mechanism in StyleInject, in part, make it possible to decouple semantic information and stylistic features during the text-to-image process. This decoupling helps avoid the direct fitting of the teacher model that might otherwise heavily shift the semantic representation of the student model. 
We show more results in~\cref{sup_fig:distill_dreamshaper,sup_fig:distill_epicphoto,sup_fig:distill_cv30}.

\begin{table*}[t]
\centering
\caption{Comparative results of distillation on SDMs. \textit{Teacher} denotes the original teacher models with $\mathbf{T}_1$ representing DreamShaper, $\mathbf{T}_2$ epiCPhotoGasm, and $\mathbf{T}_3$ Counterfeit-V3.0. For model distillates on student model SD 1.5 and ZH\_SD, performance is evaluated as U-net tuning ($\mathbf{S}$), LoRA adaptation ($\mathbf{S}$), and StyleInject adaptation ($\mathbf{S}$).}
\begin{tabular}{|l|ccc|}
\hline
 Model & CLIP-S ($\uparrow$)    & IR ($\uparrow$)  & PICK ($\uparrow$)   \\
\Xhline{2\arrayrulewidth}
 Teacher ($\mathbf{T}_1$ / $\mathbf{T}_2$ / $\mathbf{T}_3$)  & 0.8127 / 0.8165 / 0.7835 & 0.752 / 0.681 / 0.465 &  0.956 / 0.976 / 0.076 \\
\hline
SD 1.5  & 0.8133 & 0.346 & 0.113 \\
\hdashline[2.5pt/2.5pt]
U-net tune ($\mathbf{S}$) & 0.8133 / 0.8151 / 0.7955   &  0.734 / 0.623 / 0.513 & 0.985 / 0.912 / 0.399  \\
LoRA ($\mathbf{S}$) & 0.8182 / 0.8204 / 0.8037 &  0.772 / 0.671 / 0.654  & 0.991 / 0.916 / 0.516 \\
StyleInject ($\mathbf{S}$)  & \textbf{0.8198} / \textbf{0.8215} / \textbf{0.8146}  & \textbf{0.829} / \textbf{0.685} / \textbf{0.716} & \textbf{1.005} / \textbf{0.936} / \textbf{0.541} \\\hline
ZH\_SD  & 0.8072 & 0.879 & 0.703\\
\hdashline[2.5pt/2.5pt]
U-net tune ($\mathbf{S}$) & 0.7960 / 0.7837 / 0.7736 & 0.584 / 0.307 / 0.428 & 0.514 / 0.445 / 0.212 \\
LoRA ($\mathbf{S}$) & 0.8022 / 0.7934 / 0.7980 & 0.699 / 0.383 / 0.595  & 0.781 / 0.588 / 0.421 \\
StyleInject ($\mathbf{S}$) & \textbf{0.8066} / \textbf{0.8093} / \textbf{0.8022}  & \textbf{0.761} / \textbf{0.702} / \textbf{0.723} & \textbf{1.040} / \textbf{0.963} / \textbf{0.647} \\
\hline
\end{tabular}
\label{tab:distill_res}
\end{table*}

\begin{figure*}[t]
\centering
    \includegraphics[width=1\linewidth,page=1]{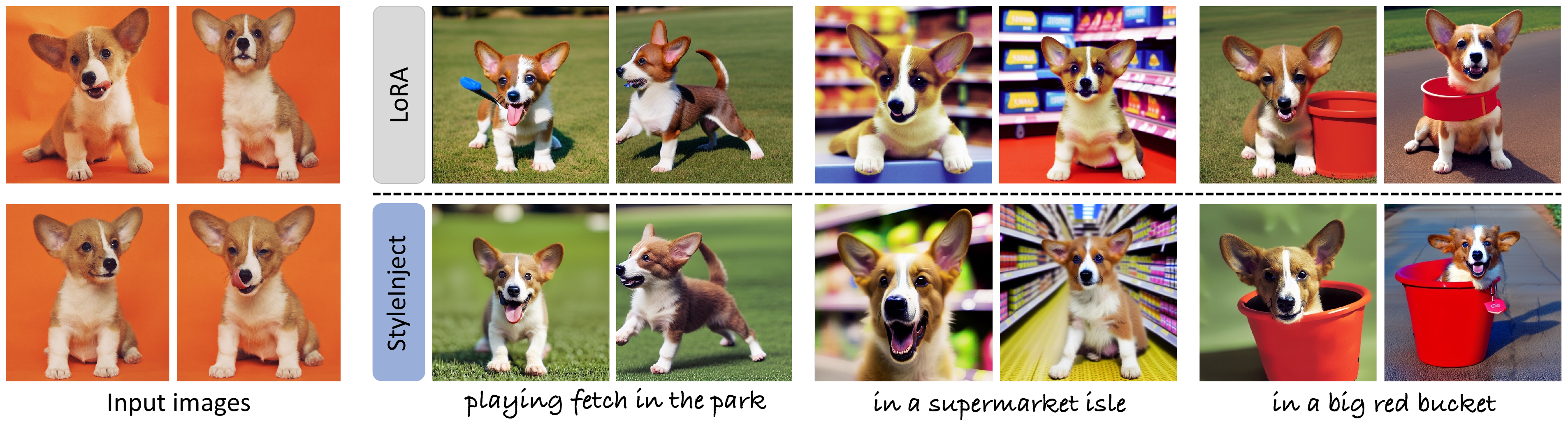}
      \caption{DreamBooth results (right) given a few images of a subject (left).}
  \label{fig:dreambooth_res}
\end{figure*}

\subsection{Subject Learning via Dreambooth}\label{sec:exp_dreambooth}
DreamBooth~\cite{ruiz2023dreambooth} is a novel approach that enables the customization generation of a specific subject with a distinct text identifier by fine-tuning a pre-trained text-to-image model using a minimal set (typically 3-5) of samples.

\cref{fig:dreambooth_res} compared the visualization results of LoRA and StyleInject for SDM model finetuning by Dreambooth. Our proposed StyleInject finetuning technique offers more natural and high-quality appearances compared to the conventional LoRA adaptation, ensuring the consistency of the subject's defining visual features. 

\section{Ablation Studies and Analysis}

In our ablation studies, we investigated the individual and combined contributions of Dynamic Multi-Style Adaptation (DMA) and Style Transfer via AdaIN (STA) to the performance of our model. We implemented DMA alone with the setting \( h^{*} = h + B\sum_{i=1}^n s_iA_i(x) \). For STA alone, we constrained the setting to \( n = 1 \). As evidenced in \cref{tab:ablation_study}, enabling either DMA or STA independently shows a positive impact on the model's performance metrics. When combined, they lead to further improvement. Notably, STA specifically boosts text-image alignment, as indicated by the increased CLIPScore, while DMA significantly enhances the overall image quality and human preference, reflected in the boost of ImageReward and PickScore metrics.

\begin{figure*}[t]
  \centering
    \includegraphics[width=1\linewidth,page=1]{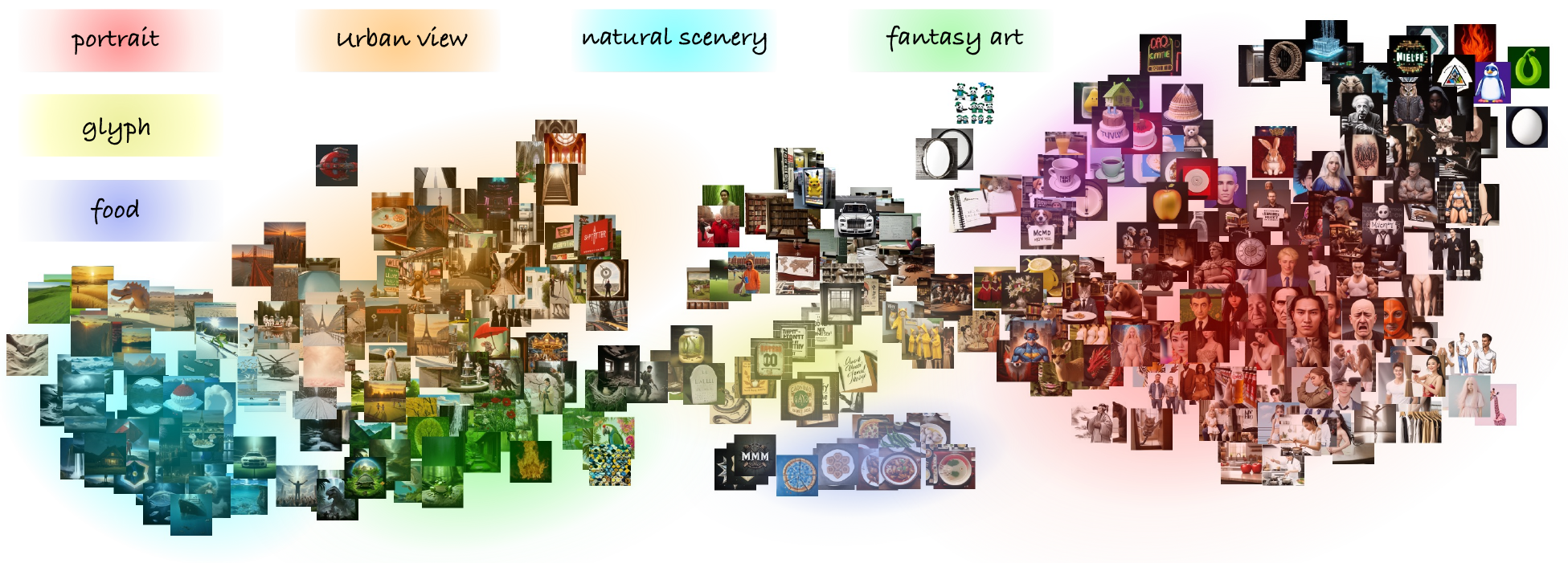}
  \caption{t-SNE visualization of combined style router outputs in StyleInject with \( n = 16 \) during inference on test prompts.}
  \label{fig:stylerouter}
\end{figure*}

\begin{figure*}[t]
  \centering
    \includegraphics[width=\linewidth,page=1]{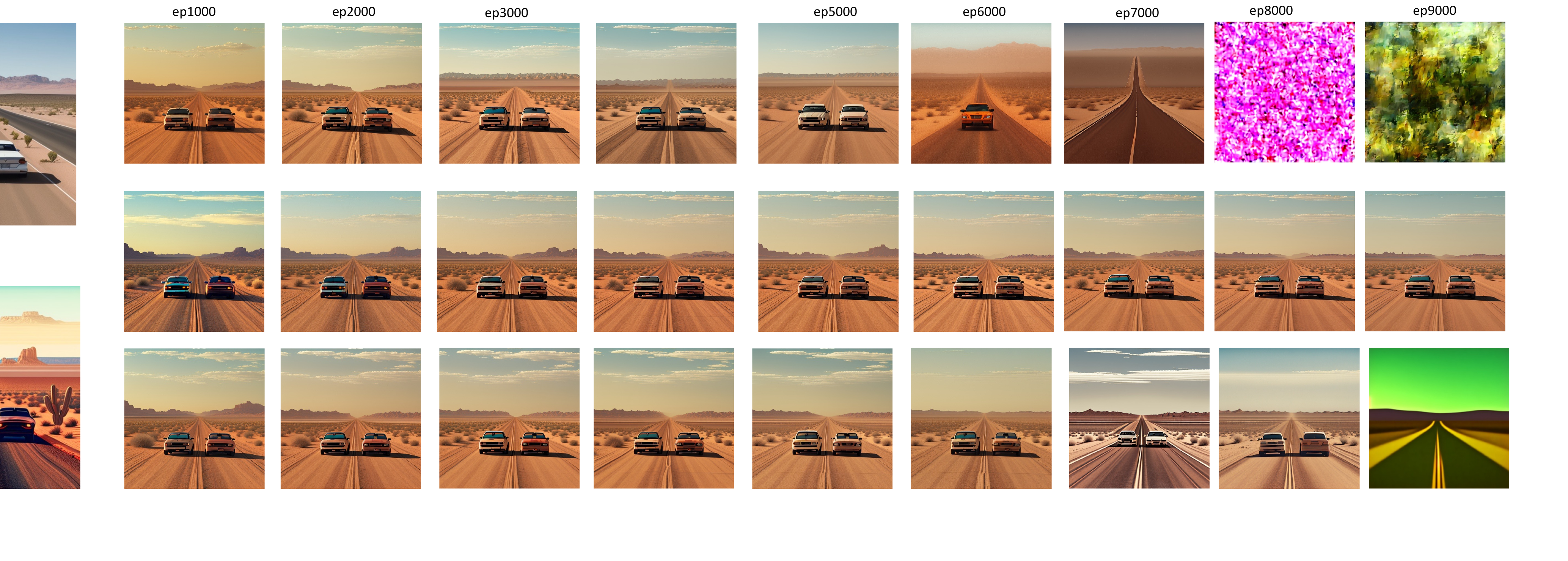}
  \caption{Examples of cross-lingual distillation process, where StyleInject is configured with \( n=1 \) and both LoRA and StyleInject have a rank size (\( r \)) of 128. Style Transformer module benefit semantic alignment.}
  \label{fig:impactSTA}
\end{figure*}

\noindent\textbf{Impact of style router}\quad We try to explore the intristic implications of the style router through the lens of t-SNE visualization~\cite{van2008visualizing}, as depicted in \cref{fig:stylerouter}, showcasing the $\textbf{s}$ cross inference denoising progress when \( n = 16 \). The visualization demonstrates the style router's capability to guide the generation of images with distinct styles. 
At the heart of its functionality lies a sophisticated mechanism capable of not only recognizing diverse visual aesthetics but also interpreting underlying semantic content embedded within images. By successfully recognizing the essence of concepts such as ``portrait'', the style router showcases its profound influence in organizing the intricate interplay between stylistic elements and conceptual semantics.
This revelation underscores the significance of the style router as a better adaptation network in navigating both the stylistic and conceptual dimensions of image generation.

\noindent\textbf{Impact of STA}\quad The cross-lingual distillation task, given its unique text encoders, poses a rigorous challenge for fine-tuning algorithms in terms of their capacity for style transfer. \cref{fig:impactSTA} reveals that during the fine-tuning process, the LoRA algorithm tends to gradually lose its precise text-image alignment capabilities as training progresses, potentially leading to model collapse. In contrast, STA, leveraging the destruction -- style modification -- construction modeling process inherent in STA, manages to mitigate the negative impact on text-image alignment, thus enabling continuous benefits in semantic alignment within a long tunning process.
This enhanced stability showcases the robustness of StyleInject's algorithm, particularly in preserving semantic coherence even when adopts to new stylistic attributes.

\begin{figure}[htbp]
  \centering
    \includegraphics[width=\linewidth,page=1]{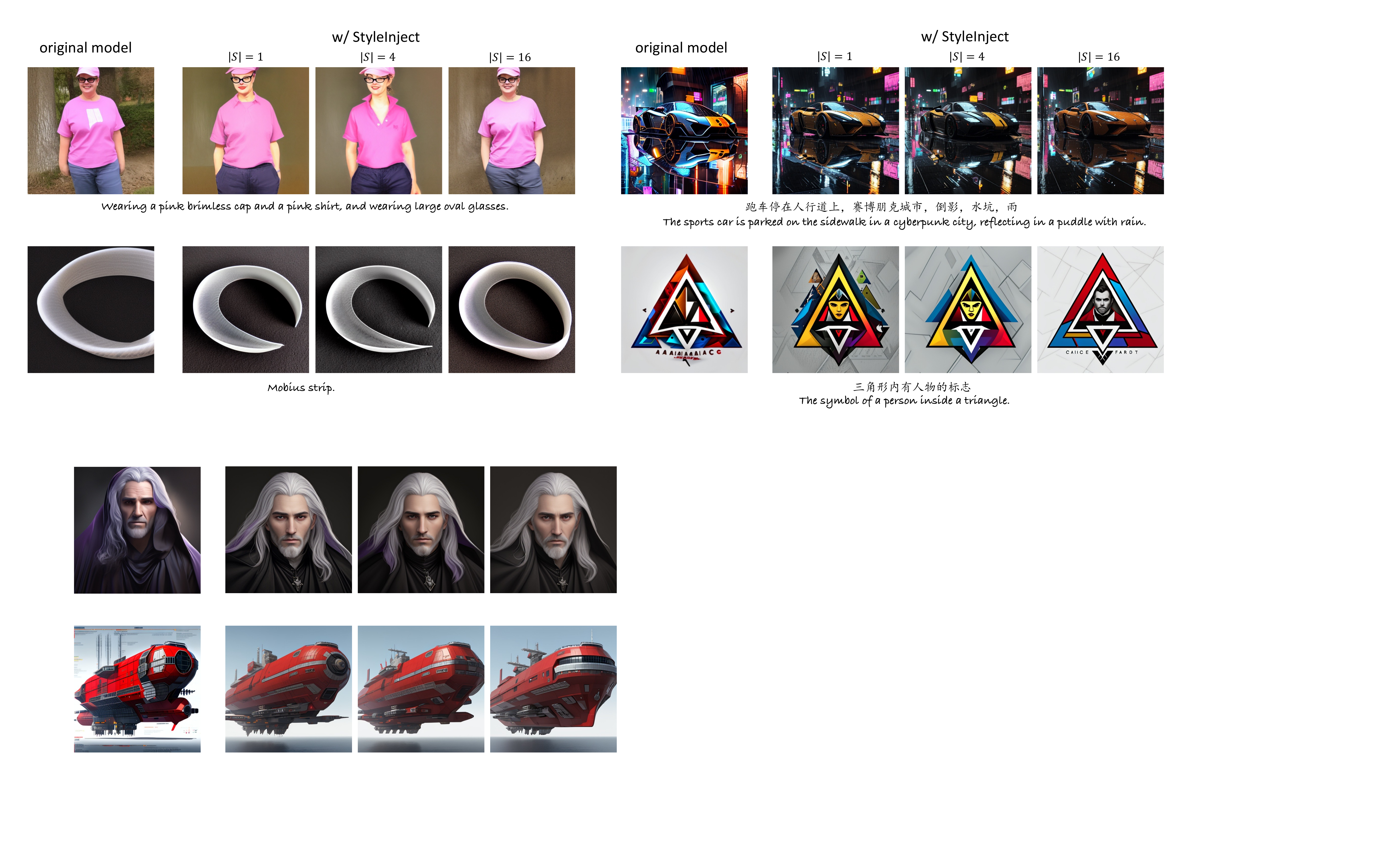}
  \caption{Enhanced quality and coherence with larger $n$.}
  \label{fig:impactN}
\end{figure}

\noindent\textbf{Impact of DMA and $n$}\quad 
The influence of incrementally adjusting $n$, while maintaining a fixed $r$, is evident from the results in \cref{tab:impactN}. Elevating $n$ leads to considerable enhancements in performance, surpassing the benefits achieved by simply increasing LoRA’s rank size $r$. As $r$ grows, StyleInject can gain steadiness improvements than the LoRA model, this also implies the superiority of our model.
The visual insights revealed by \cref{fig:impactN} further evidence this assertion, illustrating that higher values of $n$ produce richer and more informative visual outputs. Both qualitative and quantitative comparisons underscore StyleInject's scalability.
\begin{table}[t]
\renewcommand{\arraystretch}{1.05}
\centering
\caption{Comparative analysis of LoRA and StyleInject performance metrics on data-driven SD 1.5 fintuning across different settings of \( n \) and \( r \), with a consistent rank size. \#Param are denoted in millions (M).}

\begin{tabular}{|l|l|l|rccc|}
\hline
Method & $n$ & $r$ & \#Param  & CLIP-S     & IR   & PICK  \\
\Xhline{2\arrayrulewidth}
\multirow{3}{*}{LoRA} & - & 32 & 3.19 & 0.8097 & 0.387 & 0.203 \\
& - & 128 & 12.75 & 0.8093 & 0.437 & 0.197 \\
& - & 512 & 51.02 & 0.8079 & 0.443 & 0.183\\
\hline
\multirow{3}{*}{StyleInject} &1 & 32 & 3.34 & 0.8119 & 0.443 & 0.234 \\
&4 & 32 & 8.41 & 0.8135 & 0.449 & 0.240 \\
&16 & 32 & \textbf{28.70} & \textbf{0.8149} & \textbf{0.491} & \textbf{0.306} \\
\hline
\end{tabular}
\label{tab:impactN}
\end{table}

\noindent\textbf{Parameter efficiency}\quad StyleInject's utilization of a shared \( B \) matrix, coupled with the minimal parameter overhead inherent in the STA, substantiates a markedly diminished parameter footprint in comparison to LoRA at equivalent rank sizes. 
\cref{tab:impactN} shows that even as \( n \) increases and enriches the model's capability, the parameter count remains lower than what would be required for a proportional increase in LoRA's \( r \). This indicates that StyleInject not only scales well to LoRA but also with more economical use of parameters, which is an essential aspect of model design for practical application. 

\noindent\textbf{Train Epoch Settings and Early Stopping}\quad The optimal number of training epochs required to achieve the best results varies across different experimental settings. This variability is particularly noticeable in model distillation experiments involving different text encoders. As illustrated in~\cref{fig:impactSTA}, experiments using LoRA, especially with larger rank size $r$, displayed significant uncertainty, often characterized by a marked decrease in model performance or even collapse with increasing training epochs.

This deterioration can likely be attributed to the complexity and uncertainty of fitting a pixel-wise Knowledge Distillation (KD) objective function for SDMs of heterogeneous origins. A linear weight combination approach tends to simply guide the student network toward a trivial solution, essentially regressing to the mean pixel output of the teacher model, adversely affecting the overall performance of the student model. 
To address this, we employed an early stopping strategy, a common practice for LoRA training. We saved checkpoints every 1000 training steps and selected the one with the best performance for reporting the results.

\section{Conclusion}

This study introduces StyleInject, a novel adaptation approach for text-to-image diffusion models. Through an exhaustive series of experiments, including data-driven fine-tuning (\cref{sec:exp_datatune}), distillation from community-based SDMs (\cref{sec:exp_distillation}), and concept learning from limited samples (\cref{sec:exp_dreambooth}), we demonstrate that StyleInject surpasses conventional fine-tuning methods such as LoRA. Noteworthy advantages include its more efficient parameter utilization, enhanced visual outcomes, and superior alignment between text input and generated images, culminating in elevated human preferences.

\bibliographystyle{IEEEtran}
\bibliography{egbib}

\begin{thebibliography}{10}
\providecommand{\url}[1]{#1}
\csname url@samestyle\endcsname
\providecommand{\newblock}{\relax}
\providecommand{\bibinfo}[2]{#2}
\providecommand{\BIBentrySTDinterwordspacing}{\spaceskip=0pt\relax}
\providecommand{\BIBentryALTinterwordstretchfactor}{4}
\providecommand{\BIBentryALTinterwordspacing}{\spaceskip=\fontdimen2\font plus
\BIBentryALTinterwordstretchfactor\fontdimen3\font minus \fontdimen4\font\relax}
\providecommand{\BIBforeignlanguage}[2]{{%
\expandafter\ifx\csname l@#1\endcsname\relax
\typeout{** WARNING: IEEEtran.bst: No hyphenation pattern has been}%
\typeout{** loaded for the language `#1'. Using the pattern for}%
\typeout{** the default language instead.}%
\else
\language=\csname l@#1\endcsname
\fi
#2}}
\providecommand{\BIBdecl}{\relax}
\BIBdecl

\bibitem{schuhmann2022laion}
C.~Schuhmann, R.~Beaumont, R.~Vencu, C.~Gordon, R.~Wightman, M.~Cherti, T.~Coombes, A.~Katta, C.~Mullis, M.~Wortsman \emph{et~al.}, ``Laion-5b: An open large-scale dataset for training next generation image-text models,'' \emph{Advances in Neural Information Processing Systems}, vol.~35, pp. 25\,278--25\,294, 2022.

\bibitem{saharia2022photorealistic}
C.~Saharia, W.~Chan, S.~Saxena, L.~Li, J.~Whang, E.~L. Denton, K.~Ghasemipour, R.~Gontijo~Lopes, B.~Karagol~Ayan, T.~Salimans \emph{et~al.}, ``Photorealistic text-to-image diffusion models with deep language understanding,'' \emph{Advances in Neural Information Processing Systems}, vol.~35, pp. 36\,479--36\,494, 2022.

\bibitem{mishkin2022risks}
\BIBentryALTinterwordspacing
P.~Mishkin, L.~Ahmad, M.~Brundage, G.~Krueger, and G.~Sastry, ``Dall·e 2 preview - risks and limitations,'' 2022. [Online]. Available: \url{https://github.com/openai/dalle-2-preview/blob/main/system-card.md}
\BIBentrySTDinterwordspacing

\bibitem{rombach2022high}
R.~Rombach, A.~Blattmann, D.~Lorenz, P.~Esser, and B.~Ommer, ``High-resolution image synthesis with latent diffusion models,'' in \emph{Proceedings of the IEEE/CVF conference on computer vision and pattern recognition}, 2022, pp. 10\,684--10\,695.

\bibitem{nichol2021glide}
A.~Nichol, P.~Dhariwal, A.~Ramesh, P.~Shyam, P.~Mishkin, B.~McGrew, I.~Sutskever, and M.~Chen, ``Glide: Towards photorealistic image generation and editing with text-guided diffusion models,'' \emph{arXiv preprint arXiv:2112.10741}, 2021.

\bibitem{podell2023sdxl}
D.~Podell, Z.~English, K.~Lacey, A.~Blattmann, T.~Dockhorn, J.~M{\"u}ller, J.~Penna, and R.~Rombach, ``Sdxl: Improving latent diffusion models for high-resolution image synthesis,'' \emph{arXiv preprint arXiv:2307.01952}, 2023.

\bibitem{hu2021lora}
E.~J. Hu, Y.~Shen, P.~Wallis, Z.~Allen-Zhu, Y.~Li, S.~Wang, L.~Wang, and W.~Chen, ``Lora: Low-rank adaptation of large language models,'' \emph{arXiv preprint arXiv:2106.09685}, 2021.

\bibitem{liu2019roberta}
Y.~Liu, M.~Ott, N.~Goyal, J.~Du, M.~Joshi, D.~Chen, O.~Levy, M.~Lewis, L.~Zettlemoyer, and V.~Stoyanov, ``Roberta: A robustly optimized bert pretraining approach,'' \emph{arXiv preprint arXiv:1907.11692}, 2019.

\bibitem{brown2020language}
T.~Brown, B.~Mann, N.~Ryder, M.~Subbiah, J.~D. Kaplan, P.~Dhariwal, A.~Neelakantan, P.~Shyam, G.~Sastry, A.~Askell \emph{et~al.}, ``Language models are few-shot learners,'' \emph{Advances in neural information processing systems}, vol.~33, pp. 1877--1901, 2020.

\bibitem{huang2017arbitrary}
X.~Huang and S.~Belongie, ``Arbitrary style transfer in real-time with adaptive instance normalization,'' in \emph{Proceedings of the IEEE international conference on computer vision}, 2017, pp. 1501--1510.

\bibitem{nam2018batch}
H.~Nam and H.-E. Kim, ``Batch-instance normalization for adaptively style-invariant neural networks,'' \emph{Advances in Neural Information Processing Systems}, vol.~31, 2018.

\bibitem{karras2019style}
T.~Karras, S.~Laine, and T.~Aila, ``A style-based generator architecture for generative adversarial networks,'' in \emph{Proceedings of the IEEE/CVF conference on computer vision and pattern recognition}, 2019, pp. 4401--4410.

\bibitem{DBLP:journals/tmm/WangLWZF24}
Q.~Wang, S.~Li, Z.~Wang, X.~Zhang, and G.~Feng, ``Multi-source style transfer via style disentanglement network,'' \emph{IEEE Transactions on Multimedia}, vol.~26, pp. 1373--1383, 2024.

\bibitem{DBLP:journals/tmm/BaiXZD24}
Z.~Bai, H.~Xu, X.~Zhang, and Q.~Ding, ``Gcsanet: Arbitrary style transfer with global context self-attentional network,'' \emph{IEEE Transactions on Multimedia}, vol.~26, pp. 1407--1420, 2024.

\bibitem{ho2020denoising}
J.~Ho, A.~Jain, and P.~Abbeel, ``Denoising diffusion probabilistic models,'' \emph{Advances in neural information processing systems}, vol.~33, pp. 6840--6851, 2020.

\bibitem{song2020denoising}
J.~Song, C.~Meng, and S.~Ermon, ``Denoising diffusion implicit models,'' \emph{arXiv preprint arXiv:2010.02502}, 2020.

\bibitem{goodfellow2014generative}
I.~Goodfellow, J.~Pouget-Abadie, M.~Mirza, B.~Xu, D.~Warde-Farley, S.~Ozair, A.~Courville, and Y.~Bengio, ``Generative adversarial nets,'' \emph{Advances in neural information processing systems}, vol.~27, 2014.

\bibitem{DBLP:journals/tmm/XiaoB24}
J.~Xiao and X.~Bi, ``Model-guided generative adversarial networks for unsupervised fine-grained image generation,'' \emph{IEEE Transactions on Multimedia}, vol.~26, pp. 1188--1199, 2024.

\bibitem{DBLP:journals/tmm/ZhengBLWYS23}
Z.~Zheng, Y.~Bin, X.~Lv, Y.~Wu, Y.~Yang, and H.~T. Shen, ``Asynchronous generative adversarial network for asymmetric unpaired image-to-image translation,'' \emph{IEEE Transactions on Multimedia}, vol.~25, pp. 2474--2487, 2023.

\bibitem{DBLP:journals/tmm/HouZLS23}
X.~Hou, X.~Zhang, Y.~Li, and L.~Shen, ``Textface: Text-to-style mapping based face generation and manipulation,'' \emph{IEEE Transactions on Multimedia}, vol.~25, pp. 3409--3419, 2023.

\bibitem{DBLP:journals/tmm/YinCP22}
J.-L. Yin, B.-H. Chen, and Y.-T. Peng, ``Two exposure fusion using prior-aware generative adversarial network,'' \emph{IEEE Transactions on Multimedia}, vol.~24, pp. 2841--2851, 2022.

\bibitem{DBLP:journals/tmm/TanLYL22}
H.~Tan, X.~Liu, B.~Yin, and X.~Li, ``Cross-modal semantic matching generative adversarial networks for text-to-image synthesis,'' \emph{IEEE Transactions on Multimedia}, vol.~24, pp. 832--845, 2021.

\bibitem{DBLP:journals/tmm/ZhangLXYLC22}
Z.~Zhang, M.~Li, H.~Xie, J.~Yu, T.~Liu, and C.~W. Chen, ``Twgan: Twin discriminator generative adversarial networks,'' \emph{IEEE Transactions on Multimedia}, vol.~24, pp. 677--688, 2022.

\bibitem{brooks2023instructpix2pix}
T.~Brooks, A.~Holynski, and A.~A. Efros, ``Instructpix2pix: Learning to follow image editing instructions,'' in \emph{Proceedings of the IEEE/CVF Conference on Computer Vision and Pattern Recognition}, 2023, pp. 18\,392--18\,402.

\bibitem{hertz2022prompt}
A.~Hertz, R.~Mokady, J.~Tenenbaum, K.~Aberman, Y.~Pritch, and D.~Cohen-Or, ``Prompt-to-prompt image editing with cross attention control,'' \emph{arXiv preprint arXiv:2208.01626}, 2022.

\bibitem{ruiz2023dreambooth}
N.~Ruiz, Y.~Li, V.~Jampani, Y.~Pritch, M.~Rubinstein, and K.~Aberman, ``Dreambooth: Fine tuning text-to-image diffusion models for subject-driven generation,'' in \emph{Proceedings of the IEEE/CVF Conference on Computer Vision and Pattern Recognition}, 2023, pp. 22\,500--22\,510.

\bibitem{Zhang_2023_ICCV}
L.~Zhang, A.~Rao, and M.~Agrawala, ``Adding conditional control to text-to-image diffusion models,'' in \emph{Proceedings of the IEEE/CVF International Conference on Computer Vision (ICCV)}, October 2023, pp. 3836--3847.

\bibitem{DBLP:journals/tmm/CaoCHZCW24}
S.~Cao, W.~Chai, S.~Hao, Y.~Zhang, H.~Chen, and G.~Wang, ``Difffashion: Reference-based fashion design with structure-aware transfer by diffusion models,'' \emph{IEEE Transactions on Multimedia}, vol.~26, pp. 3962--3975, 2024.

\bibitem{edalati2022krona}
A.~Edalati, M.~Tahaei, I.~Kobyzev, V.~P. Nia, J.~J. Clark, and M.~Rezagholizadeh, ``Krona: Parameter efficient tuning with kronecker adapter,'' \emph{arXiv preprint arXiv:2212.10650}, 2022.

\bibitem{zhang2023adaptive}
Q.~Zhang, M.~Chen, A.~Bukharin, P.~He, Y.~Cheng, W.~Chen, and T.~Zhao, ``Adaptive budget allocation for parameter-efficient fine-tuning,'' \emph{arXiv preprint arXiv:2303.10512}, 2023.

\bibitem{valipour2022dylora}
M.~Valipour, M.~Rezagholizadeh, I.~Kobyzev, and A.~Ghodsi, ``Dylora: Parameter efficient tuning of pre-trained models using dynamic search-free low-rank adaptation,'' \emph{arXiv preprint arXiv:2210.07558}, 2022.

\bibitem{meng2023distillation}
C.~Meng, R.~Rombach, R.~Gao, D.~Kingma, S.~Ermon, J.~Ho, and T.~Salimans, ``On distillation of guided diffusion models,'' in \emph{Proceedings of the IEEE/CVF Conference on Computer Vision and Pattern Recognition}, 2023, pp. 14\,297--14\,306.

\bibitem{kim2023architectural}
\BIBentryALTinterwordspacing
B.-K. Kim, H.-K. Song, T.~Castells, and S.~Choi, ``On architectural compression of text-to-image diffusion models,'' \emph{arXiv preprint arXiv:2305.15798}, 2023. [Online]. Available: \url{https://arxiv.org/abs/2305.15798}
\BIBentrySTDinterwordspacing

\bibitem{Kim_2023_ICMLW}
------, ``Bk-sdm: Architecturally compressed stable diffusion for efficient text-to-image generation,'' in \emph{Workshop on Efficient Systems for Foundation Models@ ICML2023}, 2023.

\bibitem{ha2016hypernetworks}
D.~Ha, A.~Dai, and Q.~V. Le, ``Hypernetworks,'' 2016.

\bibitem{mao2023guided}
J.~Mao, X.~Wang, and K.~Aizawa, ``Guided image synthesis via initial image editing in diffusion model,'' \emph{arXiv preprint arXiv:2305.03382}, 2023.

\bibitem{yu2023constructing}
\BIBentryALTinterwordspacing
X.~Yu, X.~Gu, H.~Liu, and J.~Sun, ``Constructing non-isotropic gaussian diffusion model using isotropic gaussian diffusion model for image editing,'' in \emph{Thirty-seventh Conference on Neural Information Processing Systems}, 2023. [Online]. Available: \url{https://openreview.net/forum?id=2Ibp83esmb}
\BIBentrySTDinterwordspacing

\bibitem{DBLP:journals/tmm/MaLXYDJ24}
X.~Ma, C.~Liu, C.~Xie, L.~Ye, Y.~Deng, and X.~Ji, ``Disjoint masking with joint distillation for efficient masked image modeling,'' \emph{IEEE Transactions on Multimedia}, 2023.

\bibitem{hessel2021clipscore}
J.~Hessel, A.~Holtzman, M.~Forbes, R.~L. Bras, and Y.~Choi, ``Clipscore: A reference-free evaluation metric for image captioning,'' \emph{arXiv preprint arXiv:2104.08718}, 2021.

\bibitem{xu2023imagereward}
J.~Xu, X.~Liu, Y.~Wu, Y.~Tong, Q.~Li, M.~Ding, J.~Tang, and Y.~Dong, ``Imagereward: Learning and evaluating human preferences for text-to-image generation,'' 2023.

\bibitem{kirstain2024pick}
Y.~Kirstain, A.~Polyak, U.~Singer, S.~Matiana, J.~Penna, and O.~Levy, ``Pick-a-pic: An open dataset of user preferences for text-to-image generation,'' \emph{Advances in Neural Information Processing Systems}, vol.~36, 2024.

\bibitem{chinese-clip}
A.~Yang, J.~Pan, J.~Lin, R.~Men, Y.~Zhang, J.~Zhou, and C.~Zhou, ``Chinese clip: Contrastive vision-language pretraining in chinese,'' \emph{arXiv preprint arXiv:2211.01335}, 2022.

\bibitem{van2008visualizing}
L.~Van~der Maaten and G.~Hinton, ``Visualizing data using t-sne.'' \emph{Journal of machine learning research}, vol.~9, no.~11, 2008.

\end{thebibliography}

\end{document}